\pdfoutput=1
\documentclass[11pt]{article}

\usepackage[]{ACL2023}

\usepackage{times}
\usepackage{latexsym}
\usepackage{graphicx}
\usepackage{algorithm}
\usepackage{algpseudocode}
\usepackage{amssymb}
\usepackage{amsmath}
\usepackage{multirow}
\usepackage{colortbl}
\usepackage{makecell}
\usepackage{subfigure} 
\usepackage{CJKutf8}
\usepackage{longtable}
\usepackage[T1]{fontenc}
\usepackage[utf8]{inputenc}

\usepackage{microtype}

\usepackage{inconsolata}

\title{Small Language Models as Effective Guides for Large Language Models in Chinese Relation Extraction}

\author{Xuemei Tang \and Jun Wang \\
        Department of Information Management, Peking University\\ 
        Research Center for Digital Humanities of Peking University\\
        \texttt{tangxuemei@stu.pku.edu.cn}}
\begin{document}
\begin{CJK*}{UTF8}{bsmi}
\maketitle
\begin{abstract}
Recently, large language models (LLMs) have been successful in relational extraction (RE) tasks, especially in the few-shot learning. An important problem in the field of RE is long-tailed data, while not much attention is paid to this problem using LLM approaches. Therefore, in this paper, we propose SLCoLM, a model collaboration framework, to mitigate the data long-tail problem. In our framework, we use the ``\textit{Training-Guide-Predict}'' strategy to combine the strengths of small pre-trained language models (SLMs) and LLMs, where a task-specific SLM framework acts as a guider, transfers task knowledge to the LLM and guides the LLM in performing RE tasks. Our experiments on an ancient Chinese RE dataset rich in relation types show that the approach facilitates RE of long-tail relation types. 
\end{abstract}

\section{Introduction}

Recently, large language models, such as GPT-3, have introduced In-context Learning (ICL) to enhance performance with few-shot learning. However, ICL is constrained by the length of the model's input, which limits the number of examples that can be used. As a result, when sufficient supervised data is available, ICL often cannot fully leverage it.
On the contrary, small pre-trained language models can quickly adapt to fit different tasks, learn task-specific knowledge, and dominate both training and reasoning when supervised data is sufficient. 
Given the advantages of SLMs and LLMs, there seem to be some complementary between them. Some studies have started to discuss the issue of SLMs enhancing the performance of LLMs~\citet{Li_Holtzman_Fried_Liang_Eisner_Hashimoto_Zettlemoyer_Lewis, Xu_Xu_Wang_Liu_Zhu_McAuley_2023} .
% use PLMs as plug-ins for LLMs, which act as a bridge between the large model and the task-specific data to achieve knowledge transfer and fusion.

In joint relation extraction, various methods like multi-task learning, span-based approaches~\cite{Eberts_Ulges_2021, Ji_Yu_Li_Ma_Wu_Tan_Liu_2020}, and sequence labeling-based methods~\cite{Wei_Su_Wang_Tian_Chang_2020} have shown promise. These methods have outstanding results in enhancing task interaction and resolving overlapping entities.
However, they often struggle with the long-tail problem due to the limited availability of labeled data for many labels. To address this issue, some studies employ data-balancing techniques, while others use carefully designed loss functions~\cite{Lin_Goyal_Girshick_He_Dollar_2020}. Recently, LLM advancement has provided new momentum for relation extraction in various domains. Although LLMs still face challenges in understanding some low-resource domains, such as ancient Chinese~\citep{Zhang_Li_2023}, they provide a novel exploration path.
% Data balancing involves resampling underrepresented data and generating more labeled instances, while model improvements include techniques like assigning different weights to each category loss, such as Focal Loss~\cite{Lin_Goyal_Girshick_He_Dollar_2020}. 
% Moreover, some approaches focus on learning a label-unknown model and mapping instances to tight regions in the embedding space. 

Motivated by the complementary strengths of SLMs and LLMs, in this paper, we propose a framework for SLM-LLM collaboration called SLCoLM, with a \textit{``Train-Guide-Predict''} strategy, that combines the strengths of SLMs and LLMs to tackle RE tasks with long-tailed problem.
Specifically, the SLM-based model is trained on the dataset to acquire task-relevant knowledge and generate initial predictions for new samples. These SLM-generated predictions are then used to guide the LLMs during the generation process. The LLMs employ a chain-of-thought (CoT) approach, providing an explanation prior to making the final prediction. Finally, we combine the outputs from both the SLMs and LLMs to obtain the final predictions.
We use an ancient Chinese RE dataset as a case study to validate the effectiveness of the proposed framework. The experimental results demonstrate that the model collaboration framework effectively alleviates the long-tail problem.

\section{Related Work}

SLMs fine-tuning with supervised data, such as BERT, RoBERTa, etc., perform well in many natural language processing tasks. However, when dealing with complex tasks (e.g., fake news recognition, machine translation, recommendation) or low-resource domain tasks, PLMs face knowledge and capacity limitations [213].
Later, researchers found that scaling-up PLMs usually enhance downstream tasks and that LLMs show some new capabilities, such as ICL, to solve complex tasks during the scaling-up process. 
However, LLMs are not used in the same way as PLMs. They guide the model through the process of processing complex tasks mainly through prompts. In addition, LLMs can make use of external tools to improve their capabilities [211]. AutoML-GPT~\cite{Zhang_Gong_Wu_Liu_Zhou_2023} generates prompts based on user requests and instructs the GPT to connect to the AI model to achieve automatic processing from data, model, and tuning, etc. Visual ChatGPT~\cite{Wu_Yin_Qi_Wang_Tang_Duan_2023} embeds a visual base model into ChatGPT to achieve multimodal generation. These works demonstrate the strong potential of LLMs to integrate external tools dedicated to improving the performance of natural language processing tasks by combining different models. 

In practice, combining the strengths of both can more comprehensively meet the needs of different tasks and scenarios. For example, models can provide knowledge to each other, ~\citet{Yang_Zhang_Yu_Bao_Wang_Wang_Xu_Ye_Xie_Chen_et_2023} proposed SuperContext to augment LLM cues by utilizing supervised knowledge provided by task-specific fine-tuning models for out-of-distribution (OOD) datasets. ~\citet{Xu_Xu_Wang_Liu_Zhu_McAuley_2023} improved on ICL to obtain SuperICL, which efficiently performs the LLM by combining it with a small plug-in model for supervisory tasks. 
~\citet{Luo_Xu_Zhao_Geng_Tao_Ma_Lin_Jiang_2023} proposed the Parametric Knowledge Guiding (PKG) framework by fine-tuning the open-source LLMs to have the ability to generate background knowledge, and then using the background knowledge as a closed-source LLMs PKG is based on open-source LLMs and allows for offline memorization of any knowledge required by LLMs. A great advantage of LLMs is the ability to generate background knowledge and explanations for the predicted results. ~\citet{Hu_Sheng_Cao_Shi_Li_Wang_Qi_2023} argued that current LLMs may not be able to replace PLMs fine-tuned in the detection of fake news, but they can be a good advisor to PLMs by providing guided explanations from multiple perspectives.

Model collaboration has also been applied in some specific tasks to mutually improve task performance. ~\citet{Hendy_Abdelrehim_Sharaf_Raunak_Gabr_Matsushita_Kim_Afify_Awadalla_2023} explored a hybrid approach that takes advantage of both GPT and Neural Machine Translation (NMT) systems, proposing the use of a Microsoft Translator (MS-Translator) as the primary system and switching to GPT as the backup system when its quality is poor. ~\citet{Cheng_Wang_Ge_Chen_Wei_Zhao_Yan_2023} proposed a synergistic framework that connects specialized translation models and generalized LLM into a single unified translation engine, which utilizes the translation results and confidence of the PLM to join the LLM context learning examples, mitigating the linguistic bias of the LLM, whilst the LLM helps mitigate the parallel data bias of PLM.
Recommend systems not only need to predict user preferences but also need to provide explanations for the recommendations, ~\citet{Luo_Cheng_Zhang_Lu_Liu_Chen_2023} proposed the LLMXRec framework, which enables the close collaboration between previous recommendation models and LLMs-based explanation generators, generating explanations while ensuring the accuracy of the recommendation models, making full use of the LLMs in generalized inference and generative capabilities. 
In terms of information extraction, 
~\citet{Ma_Cao_Hong_Sun_2023} propose ``filter-ranking'' patterns to achieve model collaboration. The PLM is used to solve the information extraction of simple samples, and the difficult samples are selected according to the PLM, allowing the large model to predict the difficult samples.

\section{Method}

% Closed-source LLMs are difficult to improve performance by fine-tuning, usually relying on strong domain knowledge and processing new tasks through ICL without much reliance on labeled data, but LLMs do not work as well as PLMs fine-tuned on many domain-specific tasks. PLMs can quickly learn task-related knowledge through fine-tuning, and usually achieve better results with domain data training, but it is difficult to achieve the desired results when there is insufficient labeled data. Therefore, LLMs and PLMs can complement each other in terms of capability, with the LLMs being able to cope with complex tasks through ICL and application of domain knowledge, while PLMs can adapt quickly and perform well in specific domains through fine-tuning.

\begin{figure*}
    \centering
    \includegraphics*[width=14cm,height=4.7cm]{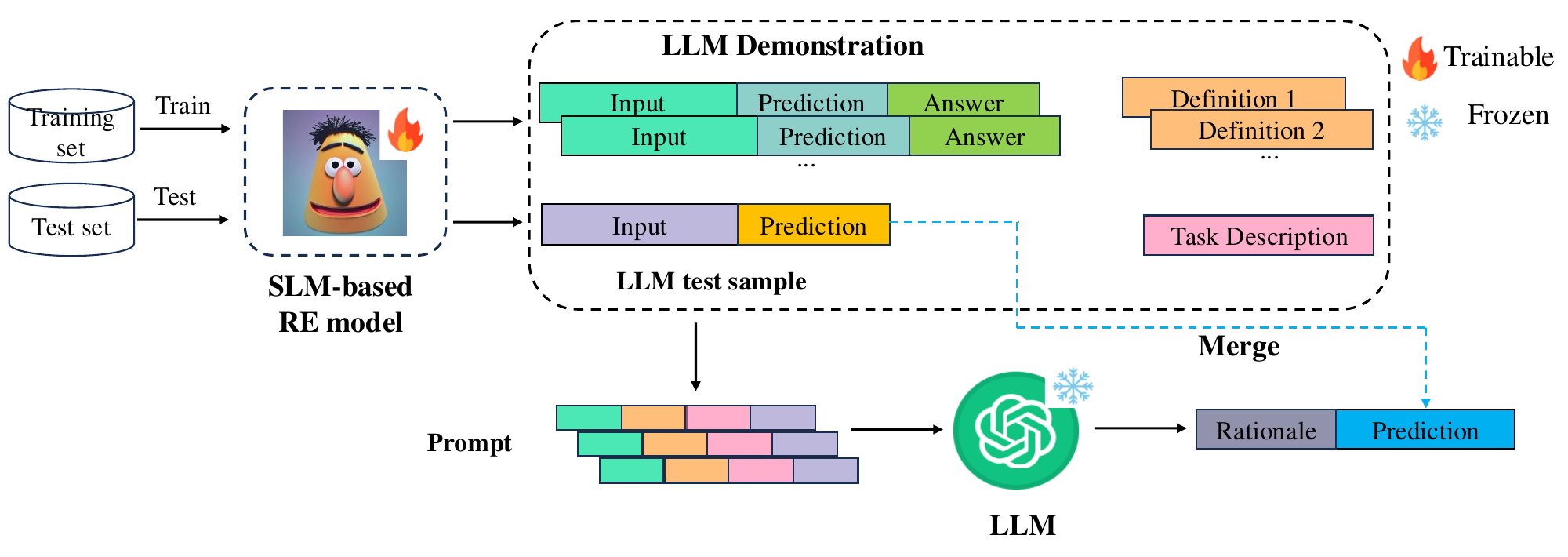}
    \caption{Model Collaboration Mechanism Illustration. ``Definition'' represents the definition of relation types.}
    \label{f1}
\end{figure*}

To alleviate the long-tailed problem, we propose a model collaboration mechanism to fully combine the advantages of SLMs and LLMs and utilize the annotated data. 
The ICL strategy of LLMs is constrained by the model's input length, which typically limits the number of demonstrations that can be included. As a result, this approach may not fully leverage the available supervised data, especially when such data is abundant. 
On the contrary, SLMs can quickly adapt to different tasks, learn task-specific knowledge, and take advantage of both training and inference when supervised data are sufficient. Given their respective strengths, there seems to be some complementarity between LLMs and SLMs. 

Therefore, we propose a ``train-guide-predict'' model collaboration mechanism, called SLCoLM, that leverages the SLMs to learn task knowledge for head categories and the LLM's extensive domain knowledge to handle long-tail categories. The illustration of model collaboration is shown in Figure~\ref{f1}.
Specifically, the SLMs are first trained with the training dataset then they learn task-related knowledge and can make predictions for new samples. Then the prediction results of SLMs are used as prompts for LLMs to transfer the task knowledge from SLMs to LLMs, and LLMs combine the task knowledge and domain knowledge to predict triples, which is not directly outputted by LLMs, but generated into an explanation, and then generates the prediction result according to the explanation. Finally, we merge the predicted results from two models to obtain the final results.

\subsection{Model Collaboration}

\begin{algorithm}[t]

    \renewcommand{\algorithmicensure}{\textbf{Input:}}
    \renewcommand{\algorithmicensure}{\textbf{Output:}} 
    \caption{Algorithm for SLCoLM.}
    \label{suanfa}
    \begin{algorithmic}[1]

    \Require Training set $D_{\text{train}}=\{(x_1,y_1),...,(x_n,y_n)\}$, validation set $D_{\text{dev}}=\{(x_1,y_1),...,(x_m,y_m)\}$, test set $D_{\text{test}}=\{(x_1,y_1),...,(x_k,y_k)\}$, LLM $M_{\text{L}}$, small SLM $M_{\text{p}}$, task description $I$, definition of candidate relations $d_c$, merge function $f_m$.
    \Ensure Predicted result ${y}_{i}$, explanation $e_i$

    \State Fine-tune SLM $M_{\text{p}}$ on training set $D_{\text{train}}$ to obtain entity relation extraction model $M_{\text{p}}^{'}$
    \State Use $M_{\text{p}}^{'}$ to predict each sample $x_i$ in $D_{\text{dev}}$ and $D_{\text{test}}$ to obtain corresponding triplet set ${y}_{i}^{*}$ and probabilities ${p}_{i}^{*}$

\For{each $x_i$ in $D_{\text{test}}$}
    \State Find $N$ similar samples from $D_{\text{dev}}$
    % $f_s(x_i,D_{\text{dev}}) \Rightarrow [(x_0,y_0,{y}_{0}^{*},{p}_{0}^{*})(x_t,y_t,\tilde{y}_{\text{plm}},\tilde{p}_{\text{plm}}),...]$ 
    \State Use $N$ samples to form demonstration set $C$
    \State \textbf{Prompt} = $C \oplus I \oplus d_c \oplus (x_i,{y}_{i}^{*},{p}_{i}^{*})$ 
    \State Use $M_{\text{L}}$ to predict ${y}_{i}^{'}$ based on \textbf{Prompt} and generate $e_i$
    \State ${y}_{i} = f_m({y}_{i}^{*},{y}_{i}^{'})$
\EndFor
\end{algorithmic}
% \caption{Algorithem for SLCoLM.}
\end{algorithm}

Within the model collaboration framework, the trained SLM-based model functions as mentors, providing their prediction results to enhance the prompts used for guiding the LLMs in predicting the input test samples. This process occurs without altering the parameters of the LLMs. By transferring knowledge from the SLM-based model to the LLMs, this approach aims to improve prediction accuracy by integrating the task-specific knowledge of the SLM-based model with the broader domain expertise of the LLMs.

The collaboration process between the SLM-based model and the LLMs is illustrated in Algorithm~\ref{suanfa}. Initially, the SLMs-based model is trained using the training set to obtain a mature RE model $M_{p}^{'}$. Subsequently, $M_{p}^{'}$ is utilized to predict all samples $x_{i}$ in the test set $D_{test}$ and validation set $D_{dev}$, generating all predicted named entities, triples form $y_{i}^{*}$ and their corresponding probabilities $p_{i}^{*}$ (lines 1-2 of the algorithm).

The next step involves finding demonstration samples for each test sample and constructing prompts for the LLM $M_{L}$(lines 4-6 of the algorithm). Firstly, for each test sample $x_{i}$ in $D_{test}$, we retrieve $N$ most semantically similar samples from the validation set $D_{dev}$, to construct a demonstration set 
$C$. The function based on cosine similarity is used to compute semantic similarity. Then, all samples are organized into the demonstration set 
$C$. As shown in Appendix ~\ref{A}, $C$ commonly used for ICL are typically direct pairings of samples and answers.
%交代为什么要使用probilities

To fully utilize the knowledge learned by the $M_{p}^{'}$, we incorporate $M_{p}^{'}$ predictions (including entities, triples, and their corresponding probabilities) as part of the demonstration, aiming to enhance the $M_{L}$ in entity recognition and relation extraction.

Next, we discuss how to achieve knowledge transfer from the SLM-based model to LLMs. 
We try to use supervised knowledge output from the SLM-based model as prompts or examples for LLM. However, the challenge of RE lies in the diversity of labeling patterns and the heterogeneity of structures~\cite{Lou_Lu_Dai_Jia_Lin_Han_Sun_Wu_2023}, so several key issues need to be noted during the collaboration process: entity relation extraction differs from the classification task in that the output of the SLM-based is not a single label or probability, but rather contains both entity and triple information. 
Therefore, we try to format the content of the demonstration as much as possible, as shown in Table~\ref{at1}, with entities and relation triples categorized by type.

After obtaining the demonstration set $C$, candidate relation type definitions 
$d_{c}$, task description $I$, and test samples $x_i$ (along with the prediction $y_i^{*}$ from the SLM-based model) are concatenated to construct the prompt.
The task description $I$ includes the task objectives and all relation labels, as shown in Table~\ref{at1}, the task objective of the LLM is to modify and supplement the predictions of the SLM-based model.
Candidate relation type definition $d_{c}$ denotes the candidate relation types' definition for the test sample $x_i$. We employ various methods to identify the most likely relation types for the test sample $x_i$, and then use the definitions of these relation types to construct $d_{c}$.\\
In section~\ref{4.1.1}, we will discuss the candidate relation type selection methods.

Finally, we use the merge function $f_m$ to combine the predicted results from the SLM-based model and the LLM (lines 8 of the algorithm). In the section ~\ref{s3.3}, we will introduce the merge function $f_m$.

\subsection{Candidate Relation Type Selection}
\label{4.1.1}

The relation types in some datasets are highly diverse, with some demonstrating strong domain-specific attributes. Therefore, providing clear and precise definitions for each relation type is crucial for enabling LLMs to extract relations accurately. We use the ancient Chinese RE dataset ChisRE as a case study, manually defining each relation type. The relation types are listed in Table~\ref{at2}, and the distribution of relation types is shown in Figure~\ref{f3}. As illustrated in Figure~\ref{f3}, this dataset exhibits a significant long-tail problem.

However, including all relation definitions in the prompt presents two challenges. First, LLMs have limited input lengths, and a lengthy prompt can significantly increase computational costs. Second, if a sentence does not involve certain relation types, including irrelevant definitions can lead to knowledge redundancy. To address these issues, we selectively add only relevant candidate relations to the prompt for each sample. We employ two methods for selecting these candidate relations, as described below.

\begin{itemize}
    \item The first method involves selecting relation types from the training set based on semantic similarity. For each test sample $x_i$, we first find the most semantically similar sample $x_j$ from the training, and then the relation types corresponding to the sample $x_j$ are included in the candidate relation set $d_c$.
    
    \item The second method is to select candidate relations based on trigger words, which are key relational representations in ancient Chinese historical texts. In most cases, they trigger some specific types of relations. For example, the ``Kill'' relation type is usually triggered by words with the semantics of ``kill'', such as ``屠殺(bloodbath)'' and ``刺殺(assassinate)''. Therefore, we first extract a list of commonly used trigger words associated with each relation type from the training set. 
    A subset of these trigger words is presented in Table~\ref{trigger}.
    Then, if a trigger word appears in the test sample, we consider it a possible trigger for a specific relation type and add it to the candidate relation set $d_c$.
\end{itemize} 
\begin{table*}
    \centering

    \begin{tabular}{c|c}
       \hline
        \Xhline{1.2pt}
         Relation Type& Trigger Words \\
          \hline
        \multirow{2}{*}{Kill} & 殺，討，謀殺，攻殺，梟，斬，襲，誅，笞殺，死，刺，刺殺，\\
        &潰，襲殺，射，活，止，屠，旣斬，斬送，收斬，射殺，策，擊殺\\
         \hline
         Affiliation & 撫，率，攻，奔，降，歸，見，得，至\\
           \hline
         Construction & 築，建，為，作，開，起 \\
          \hline
         Fearfulness & 恐，憂怖，懼 \\
          \hline
         Marriage & 出，娶，妻 \\
         \hline
    \Xhline{1.2pt}
    \end{tabular}
    \caption{Relation type trigger words (partial).}
    \label{trigger}
\end{table*}

\subsection{Merge Mode}
\label{s3.3}

In the prompt of the LLMs, although we require them to ``modify'' and ``supplement'' the results of the SLM-based model, in actual experimental processes, it was found that sometimes the LLMs only supplement the results, while other times they modify the results of the SLM-based model. Therefore, it is necessary to handle the inconsistencies between the predictions of the SLM-based model and LLMs.

To address the above problem, we adopt multiple merge methods to combine the predicted results of the LLMs and the SLM-based.
% model as the final predicted result ${y}_{i}$(line 8 of the algorithm).
Given a sample $x_i$, the set of triples predicted by the SLM-based model is denoted as $y_{i}^{*}$, and the set of predictions by the LLM is denoted as $y_{i}^{'}$. The following describes four merge methods for the final predicted results ${y}_{i}$:
\begin{itemize}
    \item Mode 1: The union of the predicted results of the SLM-based model and LLM is taken as the final predicted result: ${y}_{i} = y_{i}^{'} \cup y_{i}^{*}$.
    
    \item Mode 2: Predictions with probabilities greater than 0.6 in $y_{i}^{*}$ are selected to form $\tilde{y}_{i}^{*}$, The final result is then defined as ${y}_{i} = y_{i}^{'}\cup \tilde{y}_{i}^{*}$ representing the union of $y_{i}^{'}$ and the filtered predictions $\tilde{y}_{i}^{*}$.
 
    \item Mode 3: Remove the triple sets corresponding to the 10 relation types with the lowest F1 values from the predictions of the SLM-based model $y_{i}^{*}$, resulting in $\tilde{y}_{i}^{*}$. Subsequently, select the triple sets for these 10 relation types from LLMs predictions to form $\tilde{y}_{i}^{'}$. Finally, the final result is expressed as ${y}_{i} = \tilde{y}_{i}^{'} \cup \tilde{y}_{i}^{*}$.    
    
    \item Mode 4: Relation types are treated as discrete units. If a relation type has been predicted by the SLM-based model, all associated triples of that type predicted by the SLM-based model are added to ${y}_{i}$. Conversely, if a relation type has been predicted by LLMs but not by the SLM-base model, all corresponding triples predicted by LLMs are added to ${y}_{i}$.
    
\end{itemize}

\section{Experiments}

\subsection{Experiment Setting}

In this paper, we use an ancient Chinese relation extraction dataset ChisRE to evaluate the proposed framework. The text in the ChisRE dataset is pre-modern Chinese historical texts, including 40 relation types. The statistics of the dataset are reported in Table~\ref{at3}, and the relation type distribution is shown in Figure~\ref{f3}.

Baseline models based on PLMs are first introduced, including the generative-based model SPN~\cite{Sui_Chen_Liu_Zhao_Zeng_Liu_2020}, the sequence labeling-based model OneRel~\cite{Shang_Huang_Mao_2022}, and the span-based model Spert~\cite{Eberts_Ulges_2021}. We first compare these frameworks based on PLMs and then select the best-performing one as a PLM framework in the SLCoLM mechanism.

In the modeling collaboration framework, two key models are required, the SLM-based relation extraction model and the collaboration LLM. For the SLM-based relation extraction model, we compare a variety of commonly used entity relation joint extraction models and finally choose Spert which performs best on the dataset as the SLM-based model. 
For the collaboration LLM, we choose the closed-source GPT-3.5-turbo and ERNIE-Bot-4~\footnote{https://yiyan.baidu.com/}, which can be used by directly calling the API interface.

\subsection{Experiment Results}

\begin{table*}[t]
\centering
\setlength{\tabcolsep}{0.8mm}

\begin{tabular}{ccccccccccc} 
\hline
\Xhline{1.2pt}
\multirow{3}{*}{\textbf{Model}} & \multirow{3}{*}{\textbf{Strategy}}       & \multicolumn{3}{c|}{\textbf{NER}}                 & \multicolumn{6}{c}{\textbf{RE}}                                                                  \\ 
\cline{3-11}
                                &      & \multicolumn{3}{c|}{\textbf{Micro}}              & \multicolumn{3}{c|}{\textbf{Micro}}              & \multicolumn{3}{c}{\textbf{Macro}}            \\ 
\cline{3-11}
                             &         & \textbf{P}     & \textbf{R}     & \textbf{F}     & \textbf{P}     & \textbf{R}     & \textbf{F}     & \textbf{P}     & \textbf{R} & \textbf{F}      \\ 
\hline
\Xhline{1.2pt}
{\textbf{OneRel~\citep{Shang_Huang_Mao_2022}} }& fine-tuning & -              & -              & -              & 67.25          & 48.51          & 56.36          & 61.09          & 34.44      & 44.05           \\
{\textbf{SPN~\citep{Sui_Chen_Liu_Zhao_Zeng_Liu_2020}}}  & fine-tuning   & -              & -              & -              & 66.33          & 44.37          & 53.17          & 49.32          & 26.19      & 34.21           \\
{\textbf{Spert~\citep{Eberts_Ulges_2021}}}& fine-tuning  & \textbf{86.71} & \textbf{94.09} & \textbf{90.25} & \textbf{68.83} & 61.42          & \textbf{64.92} & 55.58          & 35.33      & 43.19           \\
\textbf{Alpaca2-7B} &LoRA                    & 85.51          & 85.28          & 85.40          & 60.68          & 39.50          & 47.85          & 42.36          & 24.07      & 30.70           \\
\hline

\multirow{4}{*}{\textbf{ERNIE-4.0}}  &Zero-shot         & 80.79          & 67.87          & 73.77          & 27.30          & 19.82          & 22.96          & 13.92          & 14.47      & 14.19           \\
& Zero-shot + SLCoLM       & \textbf{86.71} & \textbf{94.09} & \textbf{90.25}               & 60.77          & 65.53          & 63.06     &   44.45&	48.05&	46.18         \\
& ICL                & 79.90          & 66.05          & 72.32          & 41.62          & 22.81          & 29.47          & 30.59          & 16.83      & 21.72           \\
& ICL + SLCoLM            & \textbf{86.71} & \textbf{94.09} & \textbf{90.25}              & 60.34          & \textbf{66.28} & 63.17          & 48.97          & 52.80      & \textbf{50.82}  \\
\hline
\multirow{4}{*}{\textbf{GPT-3.5}}  &Zero-shot           & 74.57          & 60.74          & 66.95          & 17.29          & 18.07          & 17.67          & 9.67           & 8.25       & 8.90            \\
& Zero-shot + SLCoLM        & \textbf{86.71} & \textbf{94.09} & \textbf{90.25}               & 58.65          & 63.46          & 60.96          & 39.65          & 42.75      & 41.14           \\
& ICL               & 70.23          & 57.30          & 63.11          & 20.79          & 17.13          & 18.79          & 14.50          & 15.48      & 14.97           \\

% \textbf{SLCoLM(ERNIE, Zero-shot)}       & \textbf{86.71} & \textbf{94.09} & \textbf{90.25}               & 60.77          & 65.53          & 63.06     &   44.45&	48.05&	46.18         \\
& ICL + SLCoLM               & \textbf{86.71} & \textbf{94.09} & \textbf{90.25}              & 63.85          & 63.66          & 63.76          & \textbf{52.78} & 43.68      & 47.80           \\
\hline
\Xhline{1.2pt}
\end{tabular}
\caption{Experimental results of the SLCoLM framework on the ChisRE dataset (\%). Where ``GPT-3.5'' denotes GPT-3.5-turbo, and ``ERNIE-4.0'' denotes ERNIE-Bot-4.0; ``P'' denotes the Precision, ``R'' represents the Recall, ``F'' means the F1.}
\label{t2}
\end{table*}

The experimental results of SLCoLM on the ChisRE dataset are shown in Table~\ref{t2}. Firstly, comparing the experimental results on these SLM-based models, it can be seen that OneRel and Spert perform better relative to SPN. The Micro F1 from Spert is significantly higher than that of the other two models, and Macro F1 is slightly lower than that of OneRel. Comparing these three models, Spert has the best overall performance, and therefore it is chosen as a mentor in the SLCoLM framework.

We further analyze Spert's experimental results and observe that it predicts null values, i.e., zero F-scores, for several relation types, such as ``Motherhood'', ``Affiliation'', ``Time of Death'', ``Death Location'', and ``Former Officials''. These relation types are predominantly located in the long-tail portion of the dataset.
Observing open-source LLMs,  the fine-tuned Alpaca2 performs better on the NER task than on the RE task.
Further analysis of closed-source LLMs indicates that, in the RE task, ICL surpasses zero-shot learning in performance.
This suggests that incorporating demonstration examples can substantially enhance the prediction capabilities of LLMs. However, for the NER task, zero-shot learning proves more effective. Notably, when examples are added, GPT-3.5 experiences a marked decrease in both accuracy and recall for NER.
 
% Then we compare the performance of the SLCoLM framework and the closed-source LLM. The SLCoLM has a substantial improvement in zero-shot learning results than the LLM which only uses zero-shot learning. GPT-3.5(Zero-shot + SLCoLM) has an improvement of 32.24 percentage points in Macro F-value compared to GPT (zero-shot) and ERNIE-4.0(Zero-shot + SLCoLM) improved by 32.96 percentage points compared to ERNIE (zero-shot). When using ICL, GPT-3.5(ICL+ SLCoLM) improved by 32.82 percentage points in Macro F and 44.97 percentage points in Micro F compared to GPT-3.5 (ICL), and ERNIE-4.0(ICL+ SLCoLM) improved by 29.10 percentage points in Macro F and 43.47 percentage points in Micro F compared to ERNIE (ICL).
Next, we compare the performance of the SLCoLM framework with that of the closed-source LLMs. The SLCoLM framework demonstrates a substantial improvement in zero-shot learning compared to the LLMs using zero-shot learning alone. Specifically, GPT-3.5(Zero-shot + SLCoLM) shows a 32.24 percentage point increase in Macro F1 compared to GPT-3.5(Zero-shot), while ERNIE-4.0(Zero-shot + SLCoLM) improves by 31.99 percentage points over ERNIE (Zero-shot). In the ICL setting, GPT-3.5(ICL+ SLCoLM) achieves a 32.83 percentage point increase in Macro F1 and a 44.97 percentage point increase in Micro F-score compared to GPT-3.5(ICL). Similarly, ERNIE-4.0(ICL+ SLCoLM) shows a 29.1 percentage point improvement in Macro F1 and a 33.7 percentage point improvement in Micro F-score compared to ERNIE-4.0(ICL).

Furthermore, we compare the RE experimental results of the SLCoLM framework with those of the SLM-based model, Spert. The results show a significant improvement in Macro F1 but a slight decrease in Micro F1. This suggests that the LLM module in the SLCoLM framework enhances performance in certain relation categories that are under-represented in Spert. However, it also introduces more erroneous triples, leading to a marginal decline in overall precision.

Finally, the performance of different collaboration LLMs within the SLCoLM framework is compared. By analyzing the results of GPT-3.5(Zero-shot + SLCoLM) versus ERNIE-4.0(Zero-shot + SLCoLM) and GPT-3.5(ICL+ SLCoLM) versus ERNIE-4.0(ICL+ SLCoLM), it becomes clear that the RE task performs better when ERNIE-4.0 is used as the collaboration LLM. This improvement can likely be attributed to ERNIE’s extensive training on a large-scale Chinese corpus, which provides it with an advantage in handling ancient Chinese document tasks.

In conclusion, the experimental results demonstrate that the SLCoLM framework is more effective than other models. Integrating the task-specific knowledge of the SLMs-based model into LLM prompts is a viable approach, as it effectively guides the LLM to make more accurate predictions.

\begin{table*}
\centering
\caption{Ablation Experiment (\%). ``Dem.'' indicates demonstration examples, ``Ref.'' indicates predictions from the SLM-based model (Spert), and ``Def.'' indicates the definition of candidate relation types.}
\label{t5}
\arrayrulecolor{black}
\begin{tabular}{c|c|c|c|c|c|c|c|c|c|c} 
\hline
\Xhline{1.2pt}
\multirow{3}{*}{\textbf{Models}}  & \multirow{3}{*}{\textbf{ID}} & \multicolumn{3}{c|}{\multirow{2}{*}{\textbf{Components}}} & \multicolumn{6}{c}{\textbf{RE}}                                                                      \\ 
\cline{6-11}
                                  &                              & \multicolumn{3}{c|}{}                                     & \multicolumn{3}{c|}{\textbf{Micro}}              & \multicolumn{3}{c}{\textbf{Macro}}                \\ 
\cline{3-11}
                                  &                              &Dem.  & Ref. & Def.                                        & P              & R              & F              & P              & R              & F               \\ 
\hline
\multirow{4}{*}{\textbf{GPT-3.5}}&  1 & ×  & ×     & × &  17.29          & 18.07          & 17.67          & 9.67           & 8.25       & 8.90 \\
\cline{2-11}
& 2                           & \checkmark    & ×     & ×                                           & 20.79          & 17.13          & 18.79          & 14.50          & 15.48          & 14.97           \\ 
\cline{2-11}
                                  & 3                            & \checkmark    &\checkmark      & ×                                           & \textbf{70.16} & 54.43 & 61.30 & 54.66 & 34.25          & 42.10           \\ 
\cline{2-11}
                                  % & 3                            & \checkmark    & ×    & \checkmark                                          & 59.60          & 50.74          & 54.81          & 38.41          & 34.98          & 36.62           \\ 
\cline{2-11}
                                  & 4                            & \checkmark    & \checkmark & \checkmark  
                                   & 66.74          & \textbf{56.37}        & \textbf{61.39}          & 53.35          & 39.45          & 45.36  \\ 
\hline
\multirow{4}{*}{\textbf{ERNIE-4.0}}  
  & 5                            &×    & ×    & ×  & 27.30          & 19.82          & 22.96          & 13.92          & 14.47      & 14.19  \\
\cline{2-11}
& 6                            & \checkmark    & ×    & ×                                           & 41.62          & 22.81          & 29.47          & 30.59          & 16.83          & 21.72           \\ 
\cline{2-11}
                                  & 7                            & \checkmark   & \checkmark &  ×                                         & 64.33          & 49.48          & 55.94          & \textbf{55.85}          & 37.42          & 44.81           \\ 
\cline{2-11}
                                  % & 7                            & \checkmark    & ×    & \checkmark                                           & 62.54          & 52.37          & 57.00          & 48.71          & 40.82          & 44.42           \\ 
\cline{2-11}
                                  & 8                            & \checkmark    & \checkmark    & \checkmark                                           & 61.92          & 52.60          & 56.88          & 49.57          & \textbf{45.00} & \textbf{47.17}         \\
\hline
\Xhline{1.2pt}
\end{tabular}
\arrayrulecolor{black}
\end{table*}

\begin{table*}[t]
\centering
\caption{Experimental results of different demonstration selection methods (\%). ``Zero-shot'' indicates no demonstration, ``Random'' indicates random selection of examples, and ``Similar'' indicates selection of examples based on semantic similarity.}
\label{t6}
\arrayrulecolor{black}
\begin{tabular}{c|c|c|c|c|c|c|c} 
\hline
\Xhline{1.2pt}
\multirow{3}{*}{\textbf{Model}} & \multirow{3}{*}{\textbf{Selection}} & \multicolumn{6}{c}{\textbf{RE}}                                                      \\ 
\cline{3-8}
                                &                                     & \multicolumn{3}{c|}{\textbf{Micro}}      & \multicolumn{3}{c}{\textbf{Macro}}        \\ 
\cline{3-8}
                                &                                     & \textbf{P} & \textbf{R} & \textbf{F}     & \textbf{P} & \textbf{R} & \textbf{F}      \\ 
\hline
% Spert                           & \textbf{-}                          & 68.83      & 61.42      &\textbf{ 64.92 }         & 55.58      & 35.33      & 43.19           \\ 
% \hline
ERNIE-4.0(Zero-shot)     & -                           & 60.77      & 65.53      & 63.06        &  44.45	&48.05	&46.18         \\ 
\hline

              \multirow{2}{*}{ERNIE-4.0(ICL + SLCoLM)}                  & Random                              & 63.85      & 65.97      & \textbf{64.89} & 50.24      & 44.27      & 47.06           \\ 
\cline{2-8}
                                & Similarity                             & 60.34      & 66.28      & 63.17          & 48.97      & 52.80      & \textbf{50.82}  \\ 
\hline
  GPT-3.5(Zero-shot)     & -                                & 58.65      & 63.46      & 60.96          & 39.65      & 42.75      & 41.14           \\ 
\hline
                      \multirow{2}{*}{GPT-3.5(ICL + SLCoLM)}          & Random                              & 64.71      & 62.80      & 63.73          & 52.12      & 40.61      & 45.65           \\ 
\cline{2-8}
                                & Similarity                             & 63.85      & 63.66      & 63.76 & 52.78      & 43.68      & \textbf{47.80}  \\
\hline
\Xhline{1.2pt}
\end{tabular}
\arrayrulecolor{black}

\end{table*}

\subsection{Ablation Study}
\label{ablation}

\label{ablation}

The above analysis finds that SLCoLM outperforms individual models working in isolation. In this section, we examine the impact of collaboration, specifically the influence of various elements in LLMs prompts.
The experimental results are presented in Table~\ref{t5}, which progressively ablates the candidate relation type definitions, demonstration samples, and Spert's predictions. All the results shown in Table~\ref{t5} are raw outputs from the LLMs and have not been merged with Spert's predictions.

First, by comparing the experimental results of ID-1 and ID-2, we can observe the difference between using and without demonstrations. After incorporating demonstrations, both the Micro F1 and Macro F1 scores of GPT-3.5 improved, with a notable increase of nearly 7 percentage points in Macro F1. This suggests that demonstrations effectively enhance the performance of certain relation types that originally has low F1 scores. Additionally, comparing the results of ID-5 and ID-6, we see that for ERNIE-4.0, adding demonstrations significantly boosts the Micro F1 score, but the improvement in Macro F1 is less pronounced. This indicates that while demonstrations help ERNIE-4.0 predict more correct triplets, they do not substantially improve the performance of relation types with originally low F1 scores.
% Similar to the effect of relation definitions, examples also help ERNIE predict more correct triplets, but there is not much improvement in the performance of some originally low-F-value relation types.

Next, comparing the corresponding results of ID-2 and ID-3, as well as ID-6 and ID-7, we find that adding Spert's predictions significantly improves the performance of both GPT-3.5 and ERNIE-4.0. This demonstrates that Spert's predictions play a crucial role in guiding the LLMs within the SLcoLM framework, which has proven to be highly effective.

Additionally, by comparing the experimental results of ID-3 and ID-4, we observe that adding relation definitions leads to a significant improvement in the Macro F1 score of GPT-3.5, but a decrease in its Micro F1 score. This suggests that relation definitions help GPT-3.5 improve performance on certain specific relations, but also increase the number of incorrect triplets.
Looking at the results of ID-7 and ID-8, when ERNIE-4.0 incorporates relation definitions, both F1 scores show some improvement, with a more substantial increase in Micro F1. This indicates that relation definitions help ERNIE-4.0 predict more correct triplets, but there is less improvement in relation types that initially had low F1 scores. Overall, the impact of relation definitions on GPT-3.5 aligns with the expectations of the study, enhancing the performance on relation types that originally had low F1 scores.

In summary, different components of the prompts have varying effects on different LLMs. Spert's predictions are the most impactful, significantly improving both LLMs' performance. However, the effects of candidate relation type definitions and examples differ between the LLMs. In GPT-3.5, they primarily boost the performance of relation types with originally low F1 scores, which are likely to be long-tail relation types, and this point will be examined further in subsequent analysis. For ERNIE-4.0, the definitions and examples help predict more correct triplets overall, though these improvements may not extend to long-tail relation types.

\subsection{Effect of Demonstration Selection Method}

The analysis in Section~\ref{ablation} shows that demonstrations act as external knowledge, enhancing the learning capability of LLMs. In the SLCoLM framework, when constructing prompts for LLMs, demonstrations are selected based on semantic similarity. This section explores whether different methods of selecting demonstrations impact LLMs' performance. To investigate this, we conduct experiments comparing random selection and similarity-based selection of demonstrations, with the results presented in Table~\ref{t6}. All experimental results are the merge output of the LLM and Spert.

From the results in Table~\ref{t6}, it is clear that both random and similarity-based selection of demonstrations benefit the SLCoLM framework. However, a comparison of the two methods reveals that selecting demonstrations based on semantic similarity provides greater advantages. This finding aligns with previous research, which suggests that the performance of ICL can vary depending on the examples chosen. Therefore, selecting effective examples is crucial to fully harnessing the ICL capabilities of LLMs.

\subsection{Coverage Rate of Candidate Relation Type}

The dataset used in our experiments includes a diverse range of relation types, with their definitions incorporated into the constructed prompts to provide LLMs with the necessary knowledge for interpreting these relations. To optimize input length and minimize knowledge redundancy, candidate relation types are selectively included for each test sample. Definitions of these candidate relation types are added to the LLM’s prompts only when required.
The analysis presented in Section~\ref{ablation} demonstrates the critical role that the definitions of candidate relation types play in the effectiveness of the SLCoLM framework.

This section further validates the effectiveness of the SLCoLM framework's approach to selecting candidate relation types, specifically whether the selected candidates adequately cover the true relations in the samples.
As shown in Table~\ref{t7}, the similarity-based method identifies 44.54\% of the relation types. In contrast, the trigger words method achieves a higher coverage rate of 58.17\%, likely due to the stronger connection between trigger words and relations, whereas semantic similarity tends to reflect surface-level similarities within sentences rather than deeper relation semantics. By combining these two methods, the overall coverage rate increases to 69.43\%, indicating that most of the relation types in the samples are covered. By incorporating candidate relation definitions, we infuse domain-specific knowledge into the LLMs, and effectively manage the prompt length, thereby reducing experimental costs.

\begin{table}[]
    \centering
    \caption{Coverage Rate (\%) of different ways for selecting candidate relation types.}
    \begin{tabular}{c|c}
    \hline
    % \Xhline{1.2pt}
         Selection Mode& Cover. Rate \\
         \hline
         Similarity&44.54\\
         Heuristic & 58.17\\
         Overall & 69.43 \\
    \hline
    % \Xhline{1.2pt}     
    \end{tabular}

    \label{t7}
\end{table}

\begin{figure*}
    \centering
    % 第一行
    \subfigure[\label{p1-1}] {
        \includegraphics[width=0.45\textwidth]{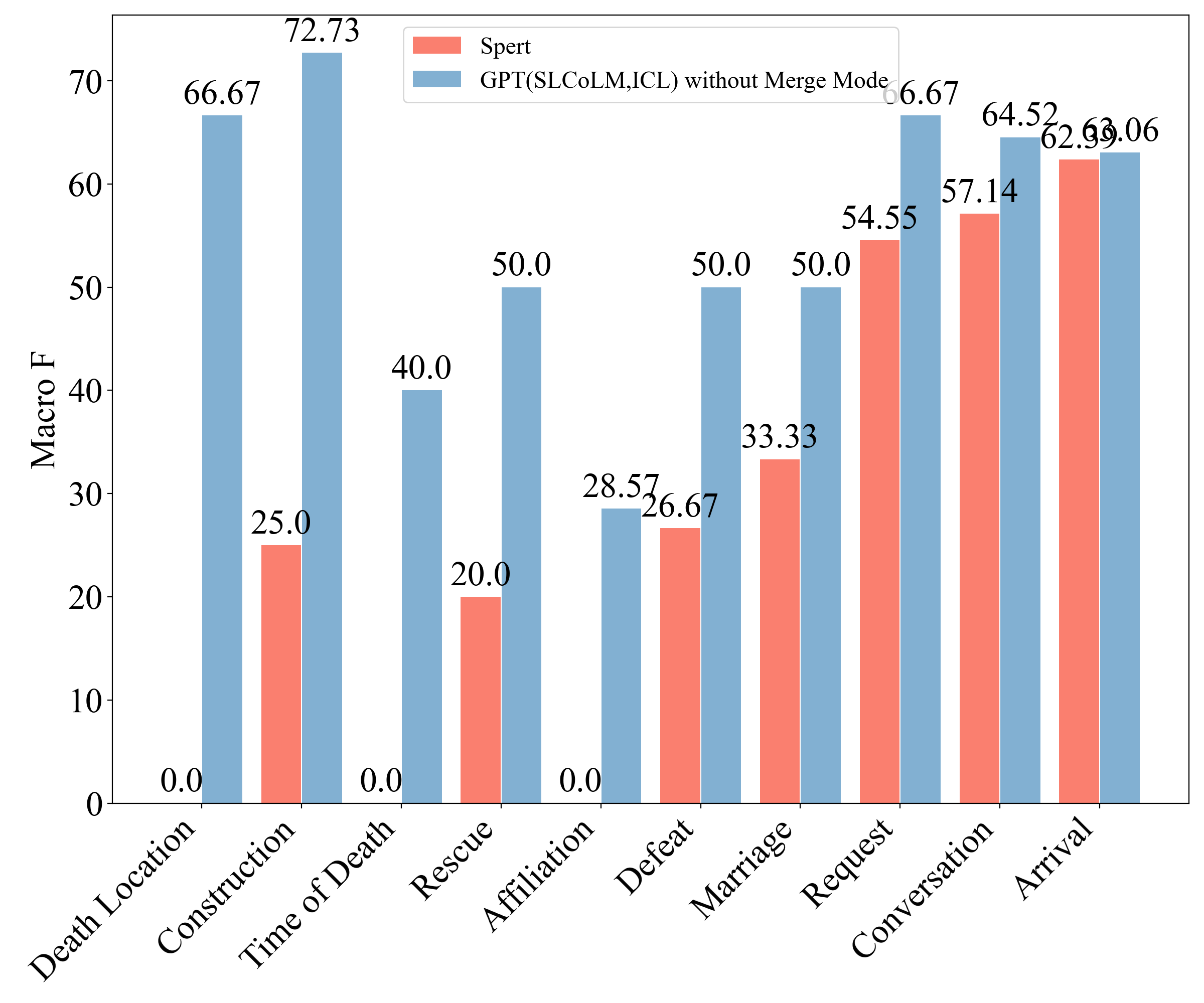}
    }
    \hfill
    \subfigure[\label{p1-2}]{
        \includegraphics[width=0.45\textwidth]{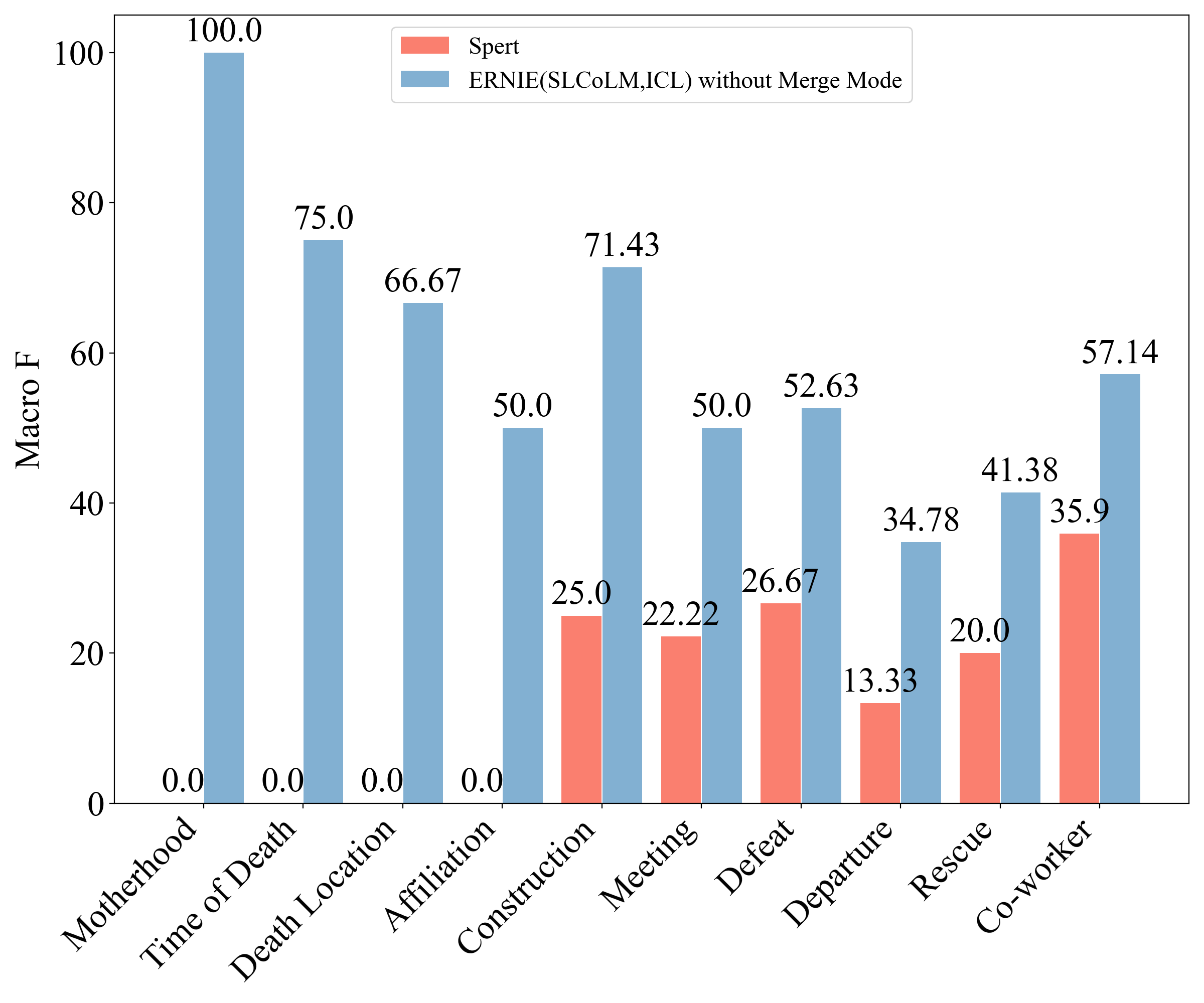}
    }
    
    % 第二行
    \vspace{1em} % 调整行间距
    \subfigure[\label{p1-3}]{
        \includegraphics[width=0.45\textwidth]{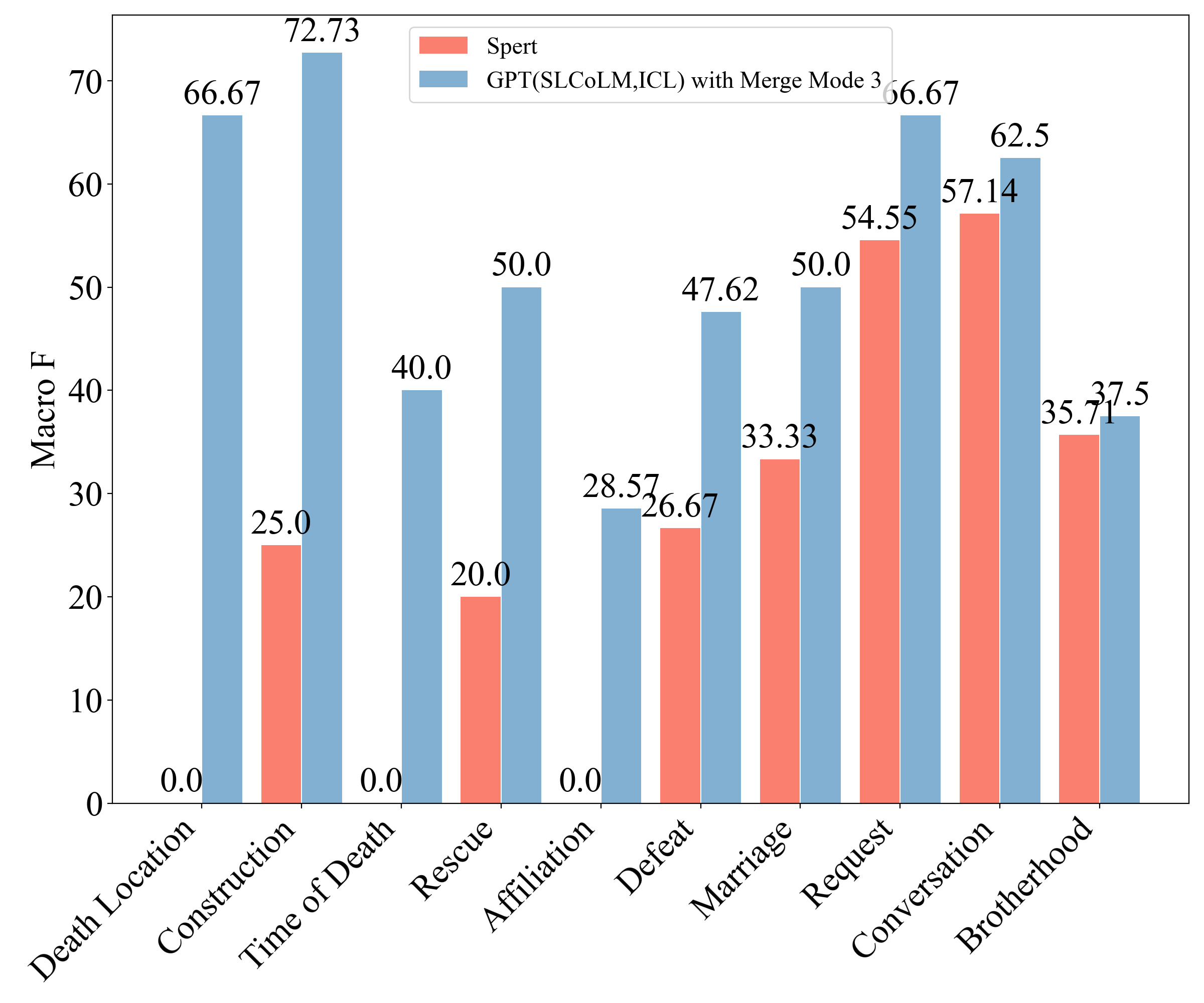}
    }
    \hfill
    \subfigure[\label{p1-4}]{
        \includegraphics[width=0.45\textwidth]{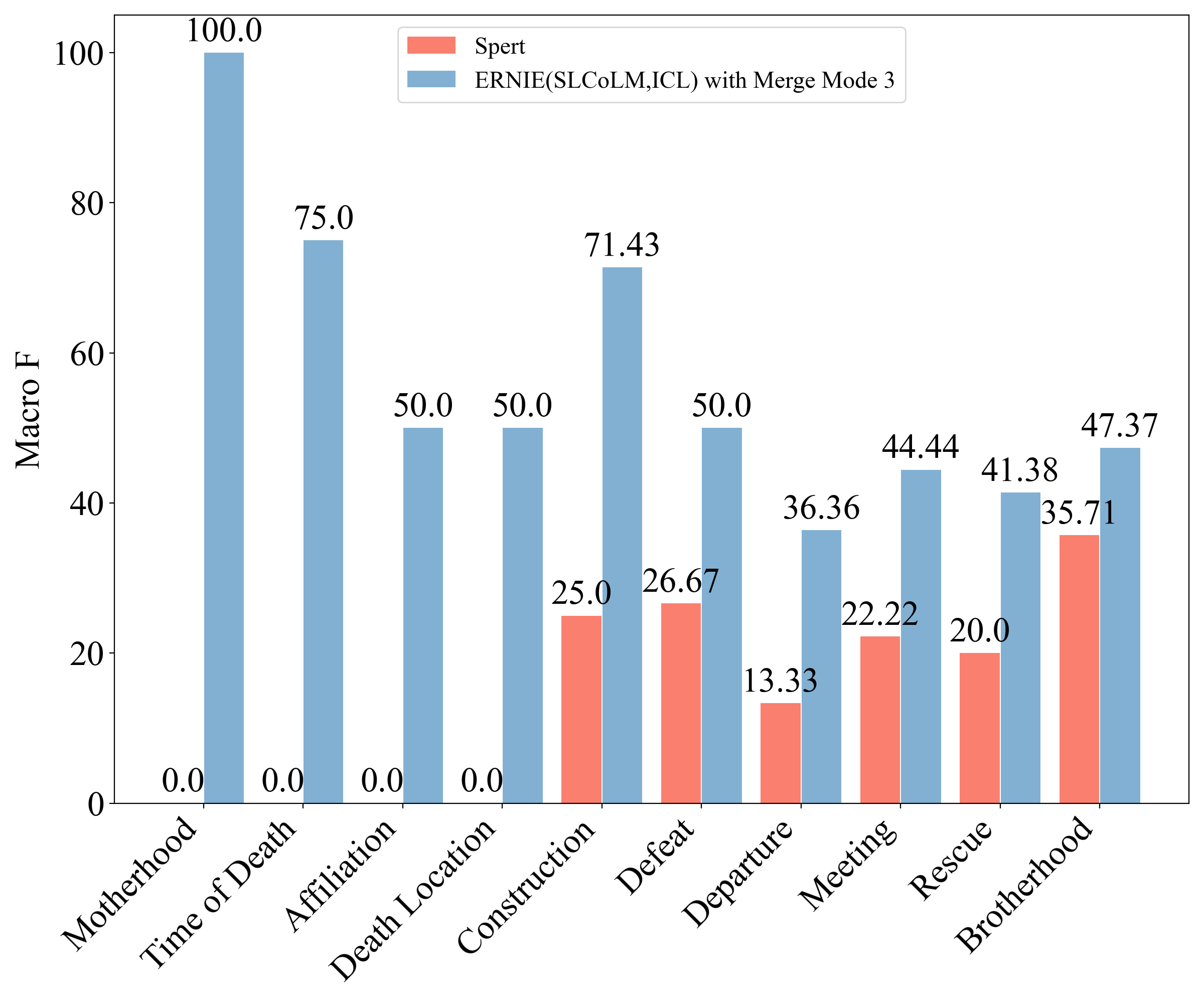}
    }
    
    \vspace{2em}
    \caption{Comparison of experimental results between the Spert and the LLMs in the SLCoLM framework (a, b), and comparison of experimental results after fusion of the Spert with the LLM in the SLCoLM framework (c, d). ``GPT(SLCoLM, Zero-shot) with Merge mode 3'' denotes the fusion of results from GPT-3.5 and Spert in the SLCoLM framework using merge mode 3.}
    \label{f4}
\end{figure*}

\subsection{Effect of Merge Mode} 

To integrate the outputs of the LLMs and the SLM-based model, we employed four distinct merging strategies. This section examines the impact of these different merging methods on the final experimental results. As shown in Table~\ref{t8}, the merge method 3 produces the most effective fusion. Under this method, GPT-3.5(ICL+ SLCoLM) and ERNIE-4.0(ICL+ SLCoLM) achieve the highest Micro F1 and Macro F1 scores, with values of 63.76 and 47.80, and 63.17 and 50.82, respectively.

To further investigate the reason behind the superior fusion performance of Mode 3, a comparative analysis is conducted by plotting the unmerged LLMs results against Spert's experimental outputs, as illustrated in Figure~\ref{f4}. In the Mode 3 setting, the 10 lowest-performing relations predicted by Spert are replaced by the corresponding LLMs predictions. These relations included ``Rescue'' ``Grandparent'', ``Departure'', ``Affiliation'', ``Motherhood'', ``Affiliation'', ``Time of Death'', ``Fearfulness'', ``Former Officials'', ``Death Location''. Additionally, other relation types predicted by the LLMs but not captured in Spert’s triple predictions were retained in the final results.

Next, we examine whether the 10 relation types that Spert struggles to extract are better handled by LLMs. Figures~\ref{p1-1} and ~\ref{p1-2} display the 10 relation types where LLMs in the SLCoLM framework show the greatest improvement over Spert. Upon reviewing these relation types, it is clear that they include categories where Spert exhibits relatively weaker performance, such as ``Time of Death'', ``Death Location'', ``Affiliation'', and ``Motherhood''. This indicates that LLMs contribute significant improvements in areas where Spert performs poorly. These findings highlight the complementary strengths of the SLM-based and LLM models in the RE task.

% Next, we discuss whether these 10 classes of relations that Spert could not accurately extract are really better handled by LLMs.
% The 10 relation types where the LLMs in the SLCoLM framework showed the greatest improvement over Spert are shown in Figures~\ref{p1-1} and ~\ref{p1-2}. Upon examining the relation types where the LLMs in SLCoLM outperform Spert, several categories in which Spert exhibits relatively weaker performance are included, such as ``Time of Death'', ``Death Location", ``Affiliation'', and ``Motherhood''.This suggests that the LLM does make many improvements in the types of relationships where Spert performs poorly. These findings underscore the complementary strengths of the SLM-based model and LLM in the RE task.

%比较图的目的1）LLM是否更好地处理了长尾类型，2）这些处理更好的结果保留在最终的结果中
Figures~\ref{p1-3} and ~\ref{p1-4} present the experimental results of the two models in SLCoLM after Mode 3 fusion, highlighting the 10 relation types that showed the greatest improvement over Spert. Notably, ERNIE-4.0 exhibits substantial enhancement for several relation types that Spert failed to predict, such as ``Motherhood'', ``Time of Death'', ``Death Location'', and ``Affiliation'', with scores increasing from 0 to above 50. This demonstrates that Mode 3 effectively maintains the relation types where LLMs outperform Spert. This approach addresses Spert's limitations and yields the most optimal final results.

\begin{table*}
\centering
\arrayrulecolor{black}
\begin{tabular}{c|c|c|c|c|c|c|c} 
\hline
\Xhline{1.2pt}

\multirow{3}{*}{\textbf{Models}}           & \multirow{3}{*}{\textbf{Merge Mode}} & \multicolumn{6}{c}{\textbf{RE}}                                                                \\ 
\cline{3-8}
                                           &                                & \multicolumn{3}{c|}{\textbf{Micro}} & \multicolumn{3}{c}{\textbf{Macro}}  \\ 
\cline{3-8}
                                           &                                & \textbf{P}     & \textbf{R}     & \textbf{F}             & \textbf{P}     & \textbf{R}     & \textbf{F}              \\ 
\hline
% \textbf{Spert}                             & \textbf{-}                     & \textbf{68.83} & 61.42          & \textbf{64.92}         & \textbf{55.58} & 35.33          & 43.19                   \\ 
% \hline
\multirow{5}{*}{\textbf{GPT-3.5(ICL+ SLCoLM)}}    & \textbf{0}                     & 67.41          & 56.37          & 61.39                  & 53.35          & 39.45          & 45.36                   \\ 
\cline{2-8}
                                           & \textbf{1}                     & 63.39          & 63.96          & 63.68                  & 49.96          & 43.81          & 46.68                   \\ 
\cline{2-8}
                                           & \textbf{2}                     & 66.24          & 58.45          & 62.10                  & 53.20          & 41.06          & 46.34                   \\ 
\cline{2-8}
                                           & \textbf{3}                     & 63.85          & 63.66          & \textbf{63.76}                  & 52.78          & 43.68          & 47.80                   \\ 
\cline{2-8}
                                           & \textbf{4}                     & 63.89          & 60.46          & 62.13                  & 49.62          & 40.97          & 44.88                   \\ 
\hline
\multirow{5}{*}{\textbf{ERNIE-4.0( ICL+ SLCoLM)}} & \textbf{0}                     & 61.92          & 52.60          & 56.88                  & 49.57          & 45.00          & 47.17                   \\ 
\cline{2-8}
                                           & \textbf{1}                     & 58.58          &67.22 & 62.60                  & 46.86          & 53.97 & 50.16                   \\ 
\cline{2-8}
                                           & \textbf{2}                     & 63.53          & 53.33          & 57.99                  & 49.83          & 45.79          & 47.72                   \\ 
\cline{2-8}
                                           & \textbf{3}                     & 60.34          & 66.28          & 63.17                  & 48.97          & 52.80          & \textbf{50.82}          \\ 
\cline{2-8}
                                           & \textbf{4}                     & 60.26          & 59.91          & 60.09                  & 46.38          & 47.21          & 46.79                   \\
\hline
\Xhline{1.2pt}
\end{tabular}
\arrayrulecolor{black}
\caption{Experimental results of different merge modes for LLM and SLM-based model results in the SLCoLM framework, ``0'' represents the preliminary results of LLM in the SLCoLM framework, while ``1'', ``2'', ``3'', and ``4'' denote the four merging modes described in Section~\ref{s3.3}.}
\label{t8}
\end{table*}

\section{Conclusion}
In this paper, we propose a model collaboration approach to address the long-tail problem in relation datasets, as well as the challenges brought by diverse relation types in domain datasets. In the model collaboration mechanism, a combination of a LLM and a SLM is utilized for dataset learning through the ``\textit{Training-Guid-Predict}'' process. The SLM is effectively employed to learn the types of head relations, while the guidance from the SLM enhances the learning of tail relation types by the LLM.

\section{Limitations}
In this paper, we validate our method using only one dataset. In the future, we plan to conduct experiments on additional datasets to further verify the effectiveness of our approach.

\bibliography{custom}
\bibliographystyle{acl_natbib}

\appendix

\section{Dataset}
\label{A}
\begin{figure*}
    \centering
    \includegraphics*[width=8.5cm,height=6cm]{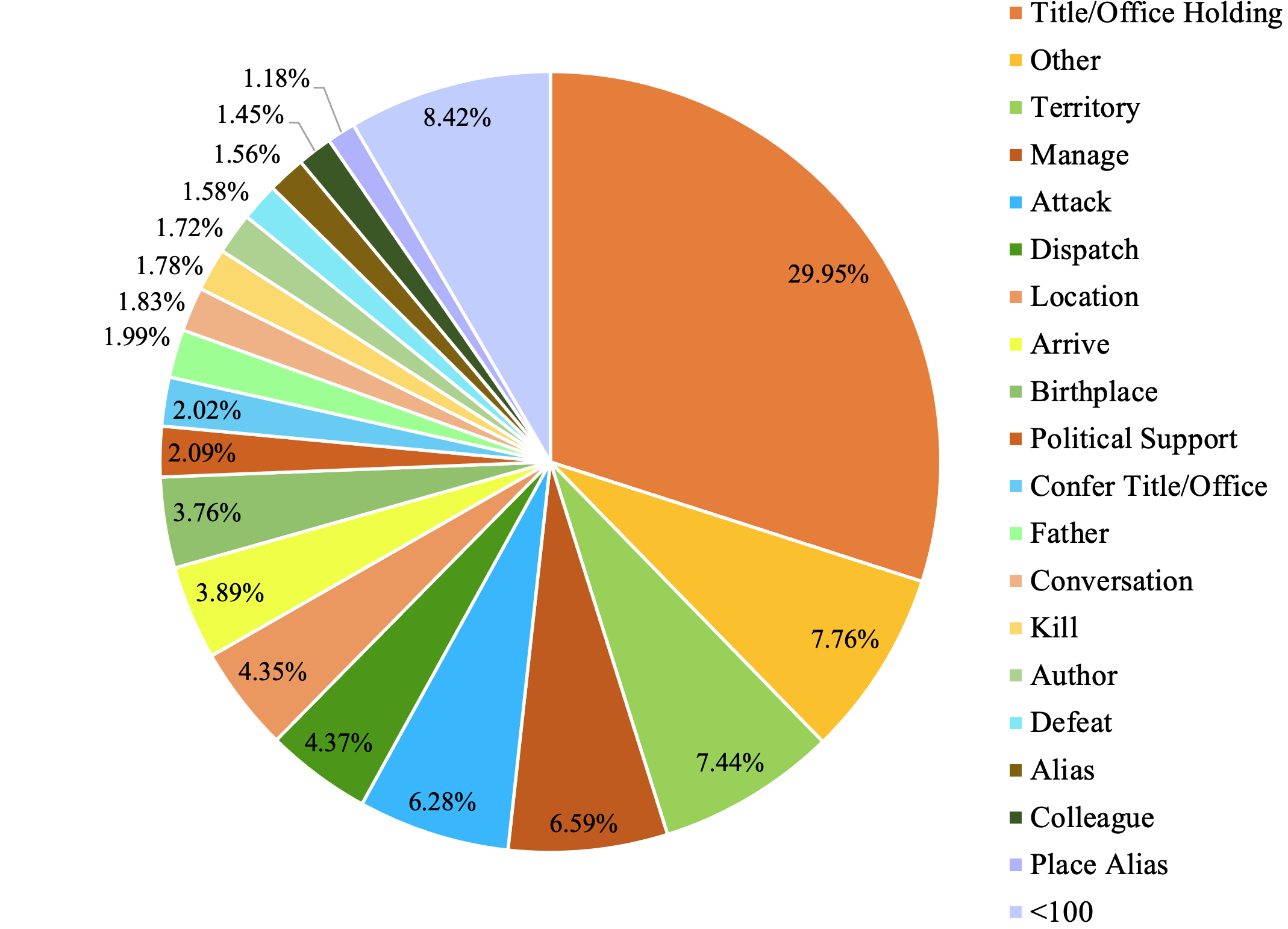}
    \caption{Percentage of each relation type. ``<100'' means these relation types with fewer than 100 samples.}
    \label{f3}
\end{figure*}

The dataset used in our experiment, ChisRE is an ancient Chinese dataset containing 6 entity types and 40 relation types (As shown is Table~\ref{at2}), and as can be seen in Figure~\ref{f3}, it is a dataset with a very serious problem of long-tailed data.
\begin{table*}
    \setlength{\tabcolsep}{0.01mm}
    \centering
    \begin{tabular}{c|c} 
    \hline
      \Xhline{1.2pt}
   \textbf{Domain} &\textbf{Relation type}\\
     \hline
   \multirow{2}{*}{Politics}& Title/Office Holding, Territory, Manage, Dispatch, Confer Title/Office, Political Support,\\
   &Colleague, Former Subordinates, Affiliation, Setup, Ally with, Former Officials\\
   \hline
    Scholar &Author \\
      \hline
   War & Attack, Kill, Defeat, Rescue \\
     \hline
   Geography &Arrive, Location, Place Alias, Departure, Pass by, Construction, Governance Location\\
     \hline
   Family & Marriage, Father, Motherhood, Brother, Grandparent \\
     \hline
   Person Information &Birthplace, Alias,  Time of Death,  Death Location \\
     \hline
   Society &Conversation, Meeting, Dependence, Collaboration, Request, Fearfulness\\
     \hline
Other & Other\\
    \hline
      \Xhline{1.2pt}
   \end{tabular}
    \caption{The history domains and the relation types.}
    \label{at2}
\end{table*}

\begin{table*}
    \centering
    \begin{tabular}{c|c|c|c} 
    \hline
      \Xhline{1.2pt}
   Dataset &Num. of Samples& Num. of Characters&Num. of Triplets \\
   \hline
   % ChisNER & 11.6K &291.1K &40.1K \\
   % \hline
   ChisRE & 4.1K & 22.3K & 9.7K\\
   \hline
   \Xhline{1.2pt}
   \end{tabular}
    \caption{Statistics of the ChisRE dataset.}
   \label{at3}
\end{table*}

\section{Prompt}
The prompt use for LLMs in SLCoLM is shown in Table~\ref{at1}.

\begin{table*}[b]
\centering
\arrayrulecolor{black}
\begin{tabular}{|c|l|} 
\hline
\multirow{32}{*}{ICL Prompt} & 
\\
&{輸入句子為：(Enter the sentence as:)}\\
&{以夏侯淵为都護將軍。}\\
&（Confer Xiahou Yuan as a general of all guards.）\\
&{模型預測的實體如下：(Model predicts entities:)}\\
&\{“人物”: [[“夏侯淵”, 0.9998]]\\
&“職官”: [[“都護將軍”, 0.9997]]\} \\
&模型預測的關係如下：(Model predicts relations:)\\
&\{“任職”: [[[“夏侯渊”, “都護將軍”], 0.9976]]\}\\
&{真實的實體：(True entities:)}\\
&{\{“人物”: [“夏侯淵”], “職官”: [“都護將軍”]\}}\\
&{真實的關係： (True relations:)}\\
&{\{“任職”: [[“夏侯淵”, “都護將軍”]]\}}\\

&{給你一些關係類型解釋： (Give you some relation type definitions:)}\\
&任職是指人物承擔了某官職，或者得到封號或者諡號。\\
&(Title/Office Holding is a person who takes an official. \\
&...\\
&現在請你在模型預測的結果上進行修改和補充，根據以上示例和關係定義，\\
&輸出以下句子中表達的真實的关系實體對，\\
&包含\textcolor{blue}{\{關係類型列表\}}。如果都不屬於以上關係，則歸屬為``其他''關係。\\
&{(Now, modify and add to the model predictions based on the above examples and relation definitions}\\
&to output the truth relation entity pair in the following input sentences.\\
&Relation type includes \textcolor{blue}{\{relation type list\}}\\
& 输入句子为：(Input sentences:)\\
&董卓以其弟旻為左將軍，兄子璜爲中軍校尉，皆典兵事，宗族內外并列朝廷。\\
&{(Dong Zhuo appointed his brother Min as a left general and his brother's son Huang as a Zhongjun }\\
&{lieutenant, both of whom were in charge of military affairs, and Dong Zhuo's family stood in the }\\
&{imperial court.)}\\
&模型預測的實體如下：(Model predicts entities:)\\
&\{“人物”: [[“旻”, 1.0], [“璜”, 1.0], [“董卓”, 0.9999]],\\
&“職官”: [[“中軍校尉”, 0.9998], [“左將軍”, 0.9996]]\}\\
&模型預測的關係如下：(Model predicts relations:)\\
&\{“任職”: [[[“旻”, “左將軍”], “0.9999”], [[“璜”, “中軍校尉”], “0.9998”]]\}\\
\hline
\end{tabular}
\caption{A prompt example (with translation).}
\label{at1}
\end{table*}

\end{CJK*}
\end{document}